\documentclass[sigconf, screen]{acmart}
\renewcommand\footnotetextcopyrightpermission[1]{} 
\usepackage[table]{xcolor}
\usepackage{tabularx}
\usepackage{booktabs}
\usepackage{multirow}
\usepackage{array} 

\AtBeginDocument{%
  }

\setcopyright{acmlicensed}
\copyrightyear{2018}
\acmYear{2018}
\acmDOI{XXXXXXX.XXXXXXX}

\acmConference[Conference acronym 'XX]{Make sure to enter the correct
  conference title from your rights confirmation email}{June 03--05,
  2018}{Woodstock, NY}

\acmISBN{978-1-4503-XXXX-X/2018/06}

\begin{document}

\title{QASA: Quality-Aware Semantic Augmentation for Robust Multimodal Sentiment Analysis}

\author{Jiazhang Liang}
\affiliation{%
  \institution{School of Computer Science, South China Normal University}
  \city{Guangzhou}
  \state{Guangdong}
  \country{China}
}

\author{Jianheng Dai}
\affiliation{%
  \institution{School of Computer Science, South China Normal University}
  \city{Guangzhou}
  \state{Guangdong}
  \country{China}
}

\author{Miaosen Luo}
\affiliation{%
  \institution{School of Computer Science, South China Normal University}
  \city{Guangzhou}
  \state{Guangdong}
  \country{China}
}

\author{Menghua Jiang}
\affiliation{%
  \institution{School of Computer Science, South China Normal University}
  \city{Guangzhou}
  \state{Guangdong}
  \country{China}
}

\author{Sijie Mai}
\authornote{Corresponding author.}
\email{sijiemai@m.scnu.edu.cn}
\affiliation{%
  \institution{School of Computer Science, South China Normal University}
  \city{Guangzhou}
  \state{Guangdong}
  \country{China}
}

\renewcommand{\shortauthors}{Trovato et al.}

\begin{abstract}
Multimodal large language models have demonstrated strong ability in capturing semantic representations for multimodal sentiment analysis. Their capacity to learn stable and generalizable multimodal features is limited, however, by the scarcity of high-quality training data. To address this, we propose QASA (Quality-Aware Semantic Augmentation), which uses diffusion models to generate augmented visual and auditory samples, thereby enlarging the training dataset and supporting multimodal learning. The generated samples can vary in quality and may exhibit cross-modal inconsistencies. To manage this, we introduce a decoupled quality-aware scoring module that assigns training weights based on the reliability of each augmented sample. This approach reduces the influence of low-quality data and contributes to more stable and robust model training. The framework combines the generative capabilities of diffusion models with the semantic reasoning of multimodal large models, providing an automated data augmentation strategy that does not require human annotation while improving generalization and robustness under limited high-quality data. Experiments on the CH-SIMS dataset show that QASA yields a relative increase of 18.0\% and 5.9\% in five-class accuracy (Acc5) and binary accuracy (Acc2), respectively, and it also outperforms existing methods on the CMU-MOSI and MUStARD benchmarks.
\end{abstract}



\maketitle

\section{Introduction}

Multimodal Sentiment Analysis (MSA) aims to infer human emotional states by jointly modeling textual, visual, and acoustic information, serving as a central task in affective computing and human-computer interaction~\cite{PORIA201798,lai2023multimodal,zhu2023multimodal,baberwal2025systematic}. Multimodal large language models (MLLMs) have demonstrated strong capabilities in cross-modal alignment for vision-language tasks and are increasingly applied to emotion understanding~\cite{Li_2025,luo2025multimodallargelanguagemodels,lian2025mer}. Despite their representational capacity, MLLMs have not achieved commensurate gains on MSA: existing benchmarks, such as CMU-MOSI~\cite{zadeh2016mosimultimodalcorpussentiment} and CH-SIMS~\cite{yu-etal-2020-ch}, remain limited in scale and costly to annotate, and under such data-constrained conditions, even powerful models tend to overfit superficial features rather than learning stable cross-modal sentiment representations~\cite{shangguan2025resourcelimitedjointmultimodalsentiment}.

Data augmentation is a common way to address this limitation. However, existing methods have constraints in the MSA setting. Standard transformations, such as random cropping or color jittering, modify only low-level statistics and provide limited semantic diversity~\cite{Shorten2019}. Augmenting the text modality, which carries the main sentiment information, through rewriting or masking can change sentiment labels due to uncontrolled semantic drift~\cite{zhang2025multimodal}. Denoising diffusion probabilistic models (DDPMs)~\cite{NEURIPS2020_4c5bcfec} generate more diverse samples, but their quality varies, and outputs often exhibit visual artifacts or lip-sync errors~\cite{trabucco2024effective}. Existing diffusion-based methods, including DRMix~\cite{wang2025drmix} and GendataAgent~\cite{li2025gendataagent}, usually mix generated samples into training without explicit quality control for cross-modal sentiment consistency, which allows low-quality samples to corrupt affective cues and disrupt alignment~\cite{huang2025robust}.

To address this challenge, we propose QASA (Quality-Aware Semantic Augmentation), a framework for robust MSA under limited high-quality data. In this framework, only the video and audio modalities are augmented, while the text modality remains unchanged. Because text carries the primary sentiment polarity, modifying it could irreversibly affect labels. QASA expands the training distribution by introducing variations in attributes that do not convey sentiment. These variations include visual style, which is augmented using FateZero~\cite{qi2023fatezero}, and speaker timbre, which is modified using Seed-VC~\cite{liu2024zero}. This design also helps prevent the model from relying on identity- or background-specific shortcuts~\cite{wang2025inversion}.

\begin{figure}[h]
\centering
\includegraphics[width=\linewidth]{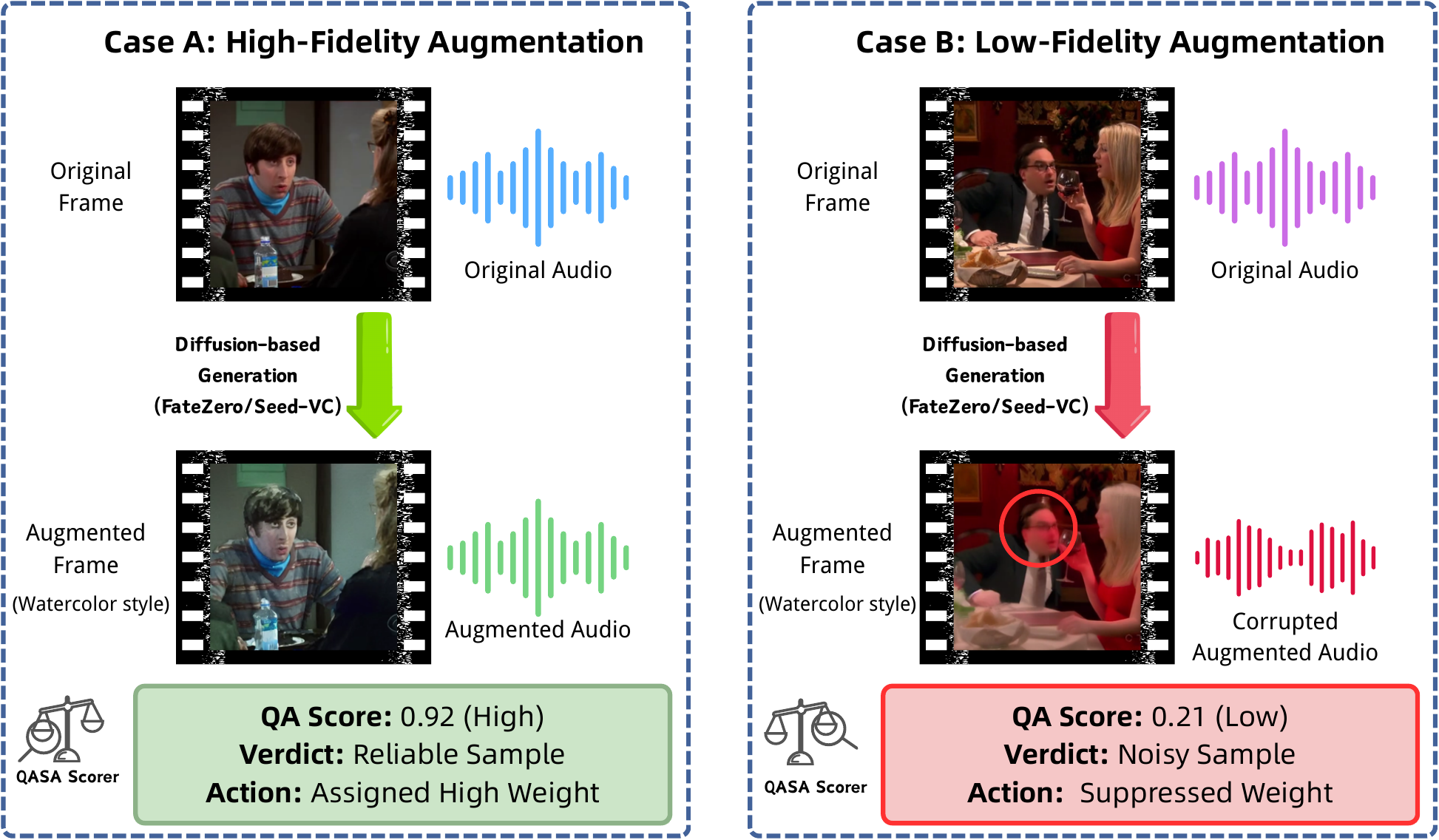}
\caption{Overview of the QASA framework. After augmenting video and audio samples via generative models, the framework assigns training weights based on Quality-Aware (QA) scores: high-fidelity samples (Case A) receive higher weights, whereas low-fidelity ones (Case B) are down-weighted.}
\label{fig:framework}
\end{figure}

Building on this approach, we note that diffusion-based generation is inherently stochastic, and the quality of the generated samples can vary. As illustrated in Figure~\ref{fig:framework}, high-fidelity augmented samples (Case A) preserve facial expressions and affective prosody and should receive high training weights. Low-fidelity samples (Case B), in contrast, exhibit facial distortion or audio corruption and should be assigned lower weights. To account for these differences, we introduce a decoupled QASA Scorer that computes a continuous quality score for each augmented sample, reflecting its preservation of cross-modal emotional consistency, and maps these scores to sample-level training weights. This generate-evaluate-weight pipeline sets QASA apart from prior diffusion-based augmentation methods by providing explicit, sentiment-aware quality control rather than mixing samples indiscriminately.

In summary, the main contributions of this paper are as follows:

\begin{itemize}
    \item We introduce QASA, a quality-aware semantic augmentation framework for MSA. The framework selectively diversifies attributes unrelated to sentiment, including visual style augmented with FateZero and speaker timbre modified with Seed-VC, while keeping the text modality unchanged. This expands the multimodal training distribution without changing sentiment labels.
    \item We design a decoupled QASA Scorer that operates entirely in the learned representation space. The scorer assigns a continuous quality score to each augmented sample and maps these scores to adaptive training weights. By performing quality control at the representation level, this method avoids the computational cost of detecting artifacts at the pixel or waveform level and effectively suppresses low-fidelity augmentations without disrupting the backbone during mixed fine-tuning.
    \item We conduct comprehensive experiments on CH-SIMS, CMU-MOSI, and MUStARD and will release all diffusion-augmented datasets, code, and model weights to support reproducibility and further research.
\end{itemize}

\section{Related Work}
\subsection{Multimodal Sentiment Analysis}
Multimodal Sentiment Analysis (MSA) focuses on understanding human emotional states during natural interactions by integrating textual, visual, and acoustic features \cite{PORIA201798,pandey2024progress, das2023multimodal, yang2025large, liu2026review}. The field has evolved from early tensor-based late fusion techniques to interactive paradigms centered on cross-modal attention mechanisms \cite{zadeh-etal-2017-tensor,pandey2024progress}. More recently, the introduction of Multimodal Large Language Models (MLLMs) has opened new avenues for affective computing, showing strong reasoning capabilities in cross-modal semantic alignment \cite{luo2025multimodallargelanguagemodels,yang2025large}. Despite their representational power, the generalization of these models remains limited by the scale and quality of available training data \cite{shangguan2025resourcelimitedjointmultimodalsentiment}.

High-quality, well-aligned multimodal sentiment datasets remain scarce. Large-scale datasets such as CMU-MOSEI contain numerous samples but were collected in unconstrained real-world settings, resulting in substantial modal variance, temporal misalignment, and subjective annotations \cite{bagher-zadeh-etal-2018-multimodal}. By contrast, datasets like CH-SIMS provide independent, fine-grained annotations for each modality and achieve higher labeling precision. However, the need for professional video materials and costly manual annotation limits the scalability of these datasets \cite{yu-etal-2020-ch}. This trade-off between scale and quality implies that existing models still lack robust support for stable feature learning \cite{han2026uncertaintyawarecollaborativelargesmall}.

Under these data-scarce conditions, models are prone to shortcut learning during optimization \cite{wan2025truth}. Empirical studies show that when cross-modal sentiment cues are insufficiently diverse, deep networks tend to rely on local confounding variables rather than capturing genuine semantic interactions. Further investigations in multimodal causal debiasing reveal that models often overfit to specific visual contexts in the training set, such as particular identities or backgrounds, introducing spurious correlations that degrade out-of-distribution performance \cite{wu-etal-2025-beyond-spurious}.

Recent work primarily addresses these feature biases through representation space interventions. For example, Meng et al. \cite{meng2026trisubspacesdisentanglementmultimodalsentiment} isolate modality-specific noise via the Tri-Subspaces Disentanglement framework to extract cleaner shared sentiment features. Although improving robustness, these approaches mainly refine existing data without fundamentally expanding the distribution or diversity of high-quality samples. A more direct solution enriches the training distribution at the source via a data augmentation strategy that reduces identity and environmental biases while preserving sentiment semantics, directly motivating our QASA framework.

\subsection{Data Augmentation and Diffusion Models}
Prior studies have explored various augmentation and generative techniques to broaden training distributions in order to address data scarcity in multimodal learning \cite{Shorten2019,alkhushayni2025multilingual}. Traditional methods typically operate on low-level statistical features and struggle to introduce high-level semantic diversity without compromising sentiment consistency \cite{mumuni2022data}. With the emergence of generative models, diffusion models have gained attention for their stable training dynamics and high-fidelity outputs, demonstrating strong augmentation potential in low-resource and imbalanced scenarios \cite{NEURIPS2020_4c5bcfec,zhang2024reconstructing}.

Recent diffusion-based methods enable temporally coherent video editing and zero-shot voice conversion, providing valuable insights for multimodal data synthesis \cite{qi2023fatezero,liu2024zero}. For instance, Trabucco et al. \cite{trabucco2024effective} proposed DA-Fusion, generating semantically diverse samples via text-to-image diffusion and reporting performance improvements in few-shot image classification. Similarly, Wang and Chen \cite{wang2025inversion} introduced Inversion Circle Interpolation to balance fidelity and diversity in data-scarce classification tasks. These cross-domain studies indicate that diffusion models can produce high-quality, varied synthetic samples, motivating the adoption of similar augmentation strategies in MSA.

Nevertheless, diffusion-generated samples often exhibit fluctuations in quality. Prior work shows that the generation process can introduce visual artifacts, prosodic distortions, or cross-modal inconsistencies, particularly in joint multimodal scenarios \cite{islam2024diffusemix,lv2025rethinking,zhao2025multimodal}. In the context of MSA, such inconsistencies can corrupt fine-grained sentiment cues and hinder cross-modal learning. Although some methods attempt to mitigate noise using unimodal priors or training dynamics, they still lack explicit verification of cross-modal semantic consistency \cite{wang2025drmix}. Moreover, most diffusion-based augmentation approaches directly mix generated samples into the training set without dedicated quality control for sentiment-sensitive tasks.

In contrast, the QASA framework addresses these limitations at the data source. It augments only the video and audio modalities, applying style transfer and timbre variation, while keeping the text modality unchanged to preserve core semantics. A decoupled quality-aware scorer constructs negative samples that simulate cross-modal inconsistency, feature degradation, and label mismatch, enabling dynamic quantification and weighting of sample reliability. This generate-evaluate-weight closed-loop mechanism leverages the semantic diversity of diffusion models and suppresses low-quality samples, leading to more stable and robust multimodal sentiment representations even with limited high-quality data.

\section{Methodology}
\subsection{Framework Overview}
To achieve robust multimodal sentiment representations under limited high-quality data, we propose the Quality-Aware Semantic Augmentation (QASA) framework. Directly mixing diffusion-generated samples into the training set inevitably introduces structural noise, such as visual artifacts or audio-visual misalignment, which can disrupt cross-modal alignment and optimization in multimodal large language models (MLLMs). To address this, QASA adopts a generate-evaluate-weight pipeline. As shown in Figure~2, the framework comprises three coordinated modules:

1) Diffusion-driven semantic augmentation (Figure~2a): To enable feature debiasing, we keep the text modality (the core carrier of sentiment semantics) unchanged and apply pretrained diffusion models to diversify video style via FateZero and timbre via Seed-VC. This design anchors the original sentiment labels while expanding the appearance distribution of the multimodal training data.

2) Feature-level quality scoring (Figures~2b--2d): As a core component of the QASA framework, we introduce a decoupled QA Scorer. Since detecting physical distortions directly at the pixel or waveform level is computationally expensive and prone to overfitting, we instead build a feature-level proxy in representation space. Concretely, we construct three types of pseudo-negative samples that simulate common failure modes of diffusion generation (cross-modal mismatch, feature corruption, and semantic polarity conflict). The QA Scorer then learns to assess cross-modal consistency and semantic fidelity of each augmented sample, outputting a continuous quality score (QA Score). The corresponding physical mappings and mathematical formulations are detailed in Section~3.3.

3) Weighted fine-tuning (Figure~2e): During fine-tuning of the backbone model (HumanOmni), we map the QA Scores to continuous sample-level training weights (the exact mapping function and fine-tuning strategy are described in Section~3.4). This decoupled weighting mechanism allows the model to suppress the influence of low-quality generated samples while learning robust cross-modal sentiment patterns from the high-fidelity augmentations.

\begin{figure*}[t]
    \centering
    \includegraphics[width=\textwidth]{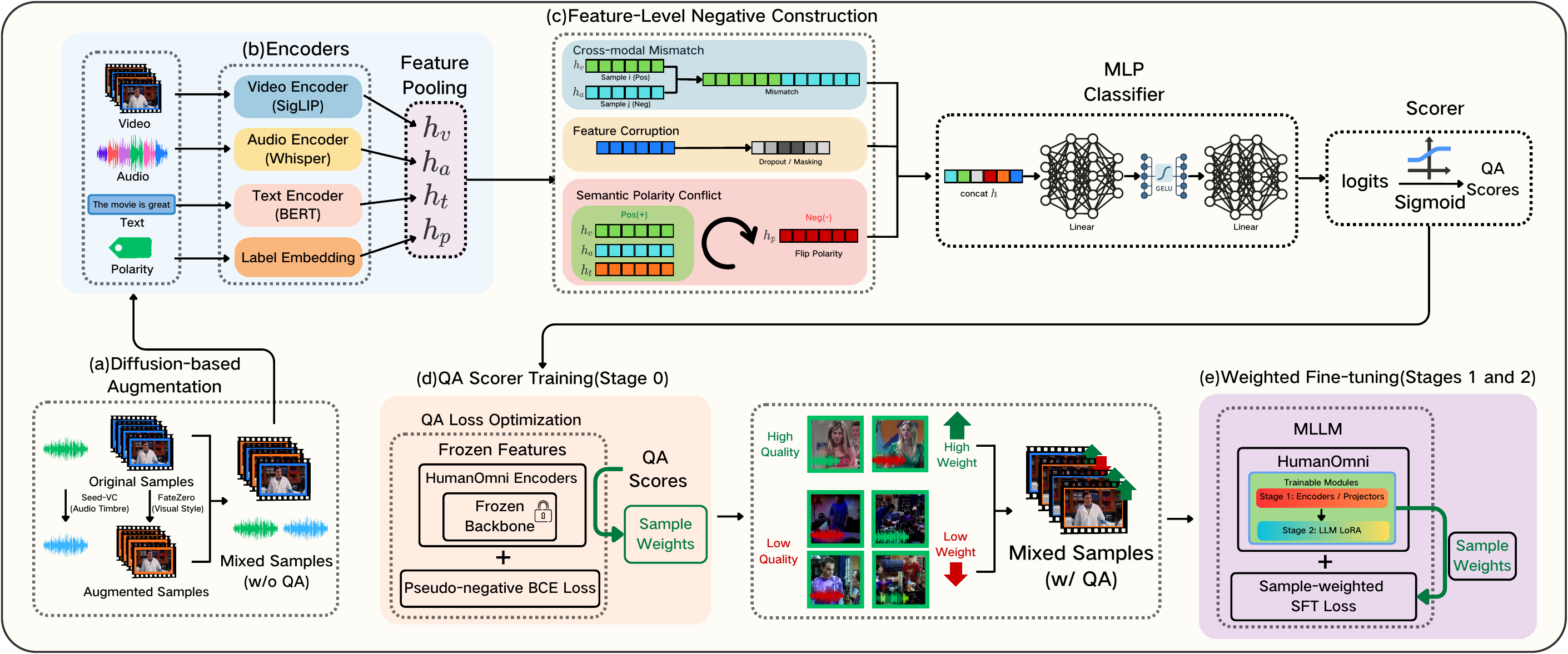}
    \caption{Overview of the QASA framework, which follows a generate-evaluate-weight pipeline. (a) Diffusion-driven augmentation of video and audio. (b) Frozen encoders extract multimodal features. (c) Feature-level negative construction. (d) Stage 0 training of the decoupled QA Scorer. (e) Stages 1 and 2 weighted fine-tuning of the HumanOmni modules using QA scores.}
    \label{fig:architecture}
\end{figure*}

\subsection{Diffusion-Based Augmentation}
\setlength{\abovedisplayskip}{3pt}
\setlength{\belowdisplayskip}{3pt}
As discussed in the introduction and related work, multimodal models with limited high-quality data are prone to shortcut learning, over-relying on spurious correlations such as specific backgrounds or speaker identities instead of intrinsic emotional cues. To mitigate this, QASA performs diffusion-based augmentation only on video and audio modalities while keeping the text modality frozen as the anchor of sentiment semantics. This selective augmentation introduces sentiment-irrelevant variations (visual style and speaker timbre) that regularize the model, encouraging it to learn more generalizable cross-modal emotional representations.

For video augmentation, we adopt FateZero~\cite{qi2023fatezero} to perform zero-shot style transfer (e.g., converting realistic scenes to watercolor styles). Standard diffusion models often distort subtle facial expressions and body movements during denoising. FateZero addresses this by introducing an attention blending mechanism. Specifically, it first performs DDIM inversion on the original video latent \(z_0\) conditioned on the source prompt \(p_{\rm src}\) to obtain the noisy latent \(z_T\) while storing the source cross- and self-attention maps:
\[
z_T = \text{DDIM-INV}(z_0, p_{\rm src}).
\]
During the subsequent editing stage at timestep \(t\), a binary mask \(M_t\) is generated from the source cross-attention maps of the edited words and used to fuse self-attention maps as
\[
M_t = \text{HEAVISIDE STEP}(c_t^{\rm src}, \tau),
\]
\[
s_t^{\rm fused} = M_t \odot s_t^{\rm edit} + (1 - M_t) \odot s_t^{\rm src},
\]
where \(\tau\) is a sensitivity threshold, \(\odot\) denotes element-wise multiplication, and this blending enforces structural and temporal consistency in non-edited regions (primarily facial muscles and body contours), enabling large stylistic changes while preserving the fine-grained dynamic cues critical for sentiment analysis.

For audio augmentation, we employ Seed-VC~\cite{liu2024zero} for zero-shot voice conversion. The goal is to replace speaker timbre while preserving prosody, pauses, and affective rhythm. Seed-VC is optimized under a flow-matching objective that aligns the predicted vector field with the true audio flow:
\[
\mathcal{L}_{\rm FM} = \mathbb{E}_{x \sim p_s,\, t \sim U[0,1]} \Bigl[ \| f_s(x,t) - v_\theta(x_t, t, c) \|_1 \Bigr],
\]
where \(c = [e_{\rm timbre}, S]\) concatenates the target timbre embedding and semantic features. To prevent timbre leakage, a random timbre perturbation is applied during training and inference:
\[
X_{\rm shifted} = \mathcal{T}(X_{\rm src}, e_r),
\]
from which clean semantic features \(S\) are extracted and injected together with the target timbre as conditioning context. This design decouples speaker identity from emotional prosody, significantly enriching acoustic diversity without altering sentiment polarity.

Collectively, these diffusion-based operations produce an augmented dataset $\mathcal{D}_{\rm aug}$ that introduces targeted variations in visual style and acoustic timbre while strictly aligning with the original sentiment labels $y_i$. Specifically, exactly one augmented sample is generated for each original training sample. In contrast to generic diffusion augmentation pipelines that indiscriminately modify all modalities or operate at the pixel/waveform level, QASA's modality-selective strategy provides a principled, sentiment-preserving expansion of the training distribution.

\subsection{Feature-Level Quality Scoring}
\setlength{\abovedisplayskip}{3pt}
\setlength{\belowdisplayskip}{3pt}
Although diffusion models provide rich semantic augmentation, their generation process is inherently stochastic and can produce samples with visual artifacts (e.g., structural facial distortion) or audio-visual desynchronization. Naively mixing such defective samples into the training set introduces structural noise that disrupts cross-modal alignment in multimodal large language models. Explicit pixel- or waveform-level artifact detectors are computationally prohibitive and prone to domain-specific overfitting.

To address this, we introduce a lightweight decoupled Quality-Aware (QA) scorer trained entirely in the learned representation space. As shown in Figure~2(b), we first extract pooled features from frozen pretrained encoders: visual features \(\mathbf{h}_v \in \mathbb{R}^d\) (SigLIP), audio features \(\mathbf{h}_a \in \mathbb{R}^d\) (Whisper), and BERT [CLS] text features that are further projected into the shared hidden space via a linear layer (\(\mathbf{h}_t \in \mathbb{R}^d\)). We additionally incorporate a binary polarity embedding \(\mathbf{h}_p \in \mathbb{R}^d\) (where label \(\geq 0\) maps to non-negative polarity and label \(< 0\) maps to negative polarity). These four components are concatenated into a unified multimodal vector:
\[
\mathbf{h} = [\mathbf{h}_v;\, \mathbf{h}_a;\, \mathbf{h}_t;\, \mathbf{h}_p] \in \mathbb{R}^{4d}.
\]

To train the QA scorer, we construct three types of feature-level pseudo-negative samples (Figure~2(c)) that simulate common diffusion failure modes. First, for cross-modal mismatch, we pair each sample \(i\) with a sample \(j\) of opposite binary polarity and randomly swap one modality (visual or audio) while keeping text and the remaining modality unchanged; this forces the scorer to detect semantic conflicts caused by audio-visual desynchronization or mismatched affective cues. Second, for feature corruption, we randomly mask feature dimensions in the pooled visual, audio, and text representations using independent Bernoulli masks, yielding corrupted features \(\tilde{\mathbf{h}}_v\), \(\tilde{\mathbf{h}}_a\), and \(\tilde{\mathbf{h}}_t\). These corrupted modality features are then concatenated with the unchanged polarity embedding \(\mathbf{h}_p\). Physical artifacts in generated videos (e.g., severe facial blurring or structural collapse) manifest in the high-dimensional latent space of pretrained encoders as the corruption or truncation of local semantic manifolds; therefore, random masking serves as a practical and effective proxy for these pixel- or waveform-level generation failures. Third, for semantic polarity conflict, we flip the polarity embedding \(\mathbf{h}_p\) while keeping all other features unchanged. This simulates catastrophic loss of the original conditioning signal.

During Stage 0 training (Figure~2(d)), the pretrained encoders are frozen and all QA inputs are detached via stop-gradient to preserve backbone stability. The scorer is parameterized as a two-layer MLP with GELU activation and is optimized using a weighted binary cross-entropy loss, treating original samples as positives (\(y=1\)) and the three pseudo-negative types as negatives (\(y=0\)):
\[
s_i = \sigma\Bigl(\mathbf{w}_2^\top \text{GELU}(\mathbf{W}_1 \mathbf{h}_i + \mathbf{b}_1) + \mathbf{b}_2\Bigr),
\]
\[
\mathcal{L}_{\rm QA} = \frac{1}{\sum_k \alpha_k} \sum_{k \in \{\rm pos,mix,mask,flip\}} \alpha_k \mathcal{L}_k,
\]
where \(\alpha_k\) are balancing hyperparameters and \(\mathcal{L}_k\) is the BCE loss for each type. The resulting QA score \(s_i \in (0,1)\) serves as a soft reliability indicator for each augmented sample.

By training exclusively on these proxy tasks in representation space, the QA scorer learns robust decision boundaries that effectively down-weight low-quality augmentations without ever operating on raw pixels or waveforms. The same representation-space QA framework is applied to all benchmarks; for sentiment analysis tasks (CH-SIMS and CMU-MOSI) we use polarity-based pseudo-negatives, while for sarcasm detection on MUStARD we adopt a task-specific variant with sarcasm-aware negative construction, including cross-sample shuffling, feature masking, noise injection, and same-class modality mixing.

\subsection{Weighted Fine-Tuning}
\setlength{\abovedisplayskip}{3pt}
\setlength{\belowdisplayskip}{3pt}
Once the QA scorer is optimized in Stage 0, we use it offline to assign quality scores to all samples and export a sample-weight file. In Stages 1 and 2 (Figure~2(e)), we fine-tune the multimodal large language model (HumanOmni) on the combined dataset \(\mathcal{D} \cup \mathcal{D}_{\rm aug}\) using a sample-level weighted optimization strategy.

To translate the QA score \(s_i \in (0,1)\) into a practical training weight, we apply a bounded power-based mapping. Original samples from \(\mathcal{D}\) receive a unit weight by default. For each augmented sample \(i\), the weight is given by
\[
w_i = w_{\rm min} + s_i^\gamma (w_{\rm max} - w_{\rm min}),
\]
where \(\gamma\) is a shaping exponent that controls the sharpness of the weight distribution, and \(w_{\rm min}\) and \(w_{\rm max}\) set the lower and upper bounds, ensuring stable training.

During supervised fine-tuning, the model generates the target response sequence \(Y_i = (y_{i,1}, \dots, y_{i,L_i})\) given the multimodal context \(X_i\). The standard autoregressive cross-entropy loss is modulated by the QA-derived sample weights:
\[
\mathcal{L}_{\rm task} = -\frac{1}{B} \sum_{i=1}^{B} w_i \left( \frac{1}{L_i} \sum_{t=1}^{L_i} \log P_\Theta(y_{i,t} \mid y_{i,<t}, X_i) \right),
\]
where \(B\) is the batch size, \(L_i\) is the effective sequence length (ignoring padding tokens marked as \texttt{IGNORE\_INDEX}), and \(\Theta\) denotes the trainable parameters in each stage: Stage 1 updates the configured tunable multimodal components, while Stage 2 further applies LoRA-based adaptation to the language model backbone.

This weighted objective ensures that high-fidelity augmented samples dominate the gradient updates while low-quality ones are significantly down-weighted. Consequently, the model expands the training distribution in a controlled manner and mitigates shortcut learning under data-scarce conditions.

\section{Experiments}
\textbf{Datasets.} 
We conduct experiments on three benchmark datasets to comprehensively evaluate the effectiveness and robustness of QASA. The Chinese multimodal sentiment analysis dataset CH-SIMS \cite{yu-etal-2020-ch} contains 2,281 video clips with five-point sentiment labels. The English multimodal sentiment analysis dataset CMU-MOSI \cite{zadeh2016mosimultimodalcorpussentiment} comprises 2,199 YouTube product review clips annotated with seven-point sentiment labels. The multimodal sarcasm detection dataset MUStARD \cite{castro2019towards} includes 690 television show clips. These datasets cover different languages and task types, providing comprehensive support for assessing QASA performance across multimodal, multilingual, and multi-task scenarios. Due to space constraints, details on baseline models, backbone, video and audio augmentation, and training pipeline are in the Supplementary Material; other information will be released with the code.

\subsection{Main Results}
Tables \ref{tab:main-results} and \ref{tab:mustard-main} evaluate the QASA framework against standard fine-tuning of mainstream multimodal large language models, including HumanOmni and Qwen2.5Omni. For fair comparison, we strictly aligned the training budget across all settings. Models on the mixed dataset (Orig. Samples $\cup$ Aug. Samples) were trained for one epoch, while baselines trained solely on either the original dataset (Orig. Samples) or the augmented dataset (Aug. Samples) were trained for two epochs. All metrics are averaged over three random seeds.

QASA achieves the best performance on both standard sentiment analysis benchmarks. As shown in Table \ref{tab:main-results}, on the CH-SIMS dataset, QASA reaches a five-class accuracy (Acc5) of 61.49\%, corresponding to an 18.0\% relative improvement over the strongest baseline HumanOmni or an absolute gain of 9.39 percentage points. The two-class accuracy (Acc2) and F1 score also increase by 5.9\% and 6.1\%, respectively. On the CMU-MOSI dataset, QASA raises Acc2 to 92.37\% and reduces MAE to 0.498. These substantial improvements, particularly in fine-grained sentiment classification (Acc5), are consistent with the proposed feature debiasing strategy.

MUStARD serves as a representative cross-modal conflict task for sarcasm detection, in which the key challenge is to capture high-frequency inconsistencies between text and audio-visual modalities. It serves as a stringent test of the robustness of QASA. As shown in Table \ref{tab:mustard-main}, QASA achieves 70.59\% on weighted accuracy (wAcc), weighted F1 (wF1), and weighted recall (wRec), representing a 4.59 percentage point improvement in wF1 over the strongest baseline Ola. This indicates that the QA scorer effectively filters low-quality generated noise that could disrupt fragile sarcasm cues. Although the weighted precision (wPrec) of 70.59 is slightly lower than the peak of 74.00 achieved by Ola, the overall improvement in F1 score demonstrates that QASA achieves a more balanced performance between precision and recall.

These results indicate that QASA consistently improves performance in both sentiment analysis and cross-modal conflict scenarios. We further investigate the sources of these performance gains through ablation studies.

\definecolor{besthl}{RGB}{230, 247, 255}

\begin{table*}[t]
\centering
\caption{Performance comparison on CH-SIMS and CMU-MOSI benchmarks. $\uparrow$ indicates higher is better; $\downarrow$ indicates lower is better. \textbf{Bold} denotes best results, \underline{underline} denotes second best.}
\label{tab:main-results}
\small
\setlength{\tabcolsep}{4.5pt}
\renewcommand{\arraystretch}{1.15}

\begin{tabular}{@{}l ccccc ccccc@{}}
\toprule
& \multicolumn{5}{c}{\textsc{CH-SIMS}} & \multicolumn{5}{c}{\textsc{CMU-MOSI}} \\
\cmidrule(lr){2-6} \cmidrule(l){7-11}
\textbf{Methods} 
& Acc5$\uparrow$ & Acc2$\uparrow$ & F1$\uparrow$ & MAE$\downarrow$ & Corr$\uparrow$
& Acc7$\uparrow$ & Acc2$\uparrow$ & F1$\uparrow$ & MAE$\downarrow$ & Corr$\uparrow$ \\
\midrule
PandaGPT\citep{luo2025multimodallargelanguagemodels}        & 38.30 & 77.20 & 74.70 & 0.431 & 0.537 & 52.10 & 90.20 & 90.20 & 0.536 & 0.899 \\
Emotion-LLaMA\citep{luo2025multimodallargelanguagemodels}   & 41.10 & 77.20 & 75.40 & 0.403 & 0.628 & 40.70 & 86.10 & 86.20 & 0.800 & 0.764 \\
MiniCPM-o\citep{luo2025multimodallargelanguagemodels}       & 48.80 & 82.50 & 80.50 & 0.350 & 0.695 & 49.80 & 89.50 & 89.50 & 0.636 & 0.853 \\
Ola\citep{luo2025multimodallargelanguagemodels}             & 48.40 & 81.60 & 80.20 & 0.406 & 0.646 & 48.30 & 89.30 & 89.30 & 0.620 & 0.860 \\
VideoLLaMA2-AV\citep{luo2025multimodallargelanguagemodels}  & \underline{52.10} & 81.60 & 82.30 & 0.388 & 0.733 & 50.40 & 90.50 & 90.50 & 0.571 & 0.877 \\
Qwen2.5Omni\citep{luo2025multimodallargelanguagemodels}     & 46.80 & 82.30 & 80.10 & \underline{0.310} & \underline{0.758} & \underline{53.90} & 90.50 & 90.50 & \underline{0.523} & \underline{0.899} \\
HumanOmni\citep{luo2025multimodallargelanguagemodels}       & \underline{52.10} & \underline{85.10} & \underline{85.00} & 0.327 & 0.749 & 52.80 & \underline{91.30} & \underline{91.30} & 0.549 & 0.881 \\
\midrule
\rowcolor{besthl}
\textbf{QASA}   & \textbf{61.49} & \textbf{90.15} & \textbf{90.18} & \textbf{0.240} & \textbf{0.832} 
                & \textbf{55.33} & \textbf{92.37} & \textbf{92.39} & \textbf{0.498} & \textbf{0.907} \\
\quad $\Delta$ \textit{vs.\ best baseline}
& \textcolor{teal}{+18.0\%} & \textcolor{teal}{+5.9\%} & \textcolor{teal}{+6.1\%} & \textcolor{teal}{-22.6\%} & \textcolor{teal}{+9.8\%}
& \textcolor{teal}{+2.7\%} & \textcolor{teal}{+1.2\%} & \textcolor{teal}{+1.2\%} & \textcolor{teal}{-4.8\%} & \textcolor{teal}{+0.9\%} \\
\bottomrule
\end{tabular}
\end{table*}

\definecolor{besthl}{RGB}{230, 247, 255}

\begin{table}[t]
\centering
\caption{Performance on the MUStARD benchmark.}
\label{tab:mustard-main}
\small
\setlength{\tabcolsep}{6pt}
\renewcommand{\arraystretch}{1.15}

\begin{tabular}{@{}l cccc@{}}
\toprule
\textbf{Methods} 
& wAcc$\uparrow$ & wF1$\uparrow$ & wPrec$\uparrow$ & wRec$\uparrow$ \\
\midrule
Ola
& \underline{67.00} & \underline{66.00} & \textbf{74.00} & \underline{67.00} \\
VideoLLaMA2
& 66.17 & 65.99 & 66.53 & 66.17 \\
Emotion‑LLaMA
& 50.00 & 44.15 & 50.00 & 50.00 \\
HumanOmni
& 66.18 & 65.57 & 67.40 & 66.18\\
\midrule
\rowcolor{besthl}
\textbf{QASA} 
& \textbf{70.59} & \textbf{70.59} & \underline{70.59} & \textbf{70.59} \\
\bottomrule
\end{tabular}
\end{table}

\subsection{Ablation Studies}
We conduct systematic ablation experiments on the CH-SIMS dataset to investigate the contribution of each component in QASA.

\subsubsection{Data Augmentation}
Table \ref{tab:ch-sims-ablation} summarizes performance under five data configurations. Training with only diffusion-augmented samples (Aug. Samples)  yields a five-class accuracy (Acc5) of 60.39\%, higher than the 58.64\% obtained with traditional augmentation methods (Traditional Aug.). Diffusion-generated samples therefore provide richer semantic diversity, whereas traditional pixel-level or statistical transformations supply only low-level variations that fail to expand the emotion-relevant distribution effectively.

Directly mixing diffusion-augmented samples with original samples at equal weights (Mixed w/o QA) further raises Acc5 to 62.58\%, although Acc2 and F1 reach only the second-best levels. The inherent quality fluctuations in the diffusion generation process introduce noise that interferes with model training. In contrast, weighting the augmented samples with the QA scorer (Mixed w/ QA) slightly lowers Acc5 to 61.49\% but raises Acc2 and F1 to 90.15\% and 90.18\%, respectively—the highest values among all configurations. The QA scorer therefore serves as an effective quality control mechanism that mitigates the impact of low-quality samples and enables more stable optimization during mixed training.

\subsubsection{Modality Contributions}
Table \ref{tab:modality-ablation} analyzes the performance of different modality combinations under three training configurations on the CH-SIMS test set, where v denotes video, a audio, t text, tv text+video, ta text+audio, and va video+audio. Across all configurations, the text-video combination (tv) consistently performs best, as text carries the core sentiment semantics while video supplies complementary information.

Under the Mixed (w/ QA) configuration, the tv combination achieves the highest performance, with Acc5 of 59.96\% and Acc2 of 90.02\%, a substantial improvement over the tv combination using only original samples (Acc5 of 54.70\%). This outcome shows that visual style transfer via FateZero successfully performs text-visual debiasing, allowing the model to focus on authentic cross-style emotional cues rather than overfitting to specific backgrounds or identities in the original videos.

More importantly, in mixed training without quality control (Mixed w/o QA), the video-audio combination (va), which depends heavily on precise audiovisual alignment, suffers a significant performance drop (Acc5 decreases from 53.83\% for Original Samples to 52.52\%). This confirms that inherent audiovisual misalignment in the diffusion generation process, such as lip-sync failures, acts as structural noise and severely disrupts cross-modal feature alignment. Introducing the QA scorer (Mixed w/ QA), however, not only halts this decline but reverses it, with the va combination surpassing the original baseline to 54.49\%. The rebound offers clear evidence that the QA module accurately identifies and filters out cross-modal failure samples, confirming its indispensable role in preserving multimodal alignment.

\begin{table}[t]
\centering
\small
\caption{\textbf{Ablation study on CH-SIMS}. }
\label{tab:ch-sims-ablation}
\setlength{\tabcolsep}{4.5pt}
\renewcommand{\arraystretch}{1.15}

\begin{tabularx}{\columnwidth}{@{}X c c c c c@{}}
\toprule
\textbf{Methods} 
& Acc5$\uparrow$ & Acc2$\uparrow$ & F1$\uparrow$ & MAE$\downarrow$ & Corr$\uparrow$ \\
\midrule
Traditional Aug.   & 58.64 & 88.18 & 87.74 & 0.246 & \underline{0.837} \\
Orig. Samples   & 52.08 & 85.10 & 85.03 & 0.327 & 0.749 \\
Aug. Samples  & 60.39 & 88.40 & 88.26 & 0.256 & 0.821 \\
Mixed (w/o QA)     & \textbf{62.58} & \underline{89.06} & \underline{89.12} & \textbf{0.232} & \textbf{0.845} \\
\midrule
\rowcolor{besthl}
Mixed (w/ QA)      & \underline{61.49} & \textbf{90.15} & \textbf{90.18} & \underline{0.240} & 0.832 \\
\bottomrule
\end{tabularx}
\end{table}

\begin{table}[t]
  \centering
  \caption{Multimodal contribution analysis on CH-SIMS.}
  \label{tab:modality-ablation}
  \small
  \setlength{\tabcolsep}{6pt}
  \renewcommand{\arraystretch}{1.12}

  \begin{tabular}{@{}l c ccc@{}}
    \toprule
    \textbf{Methods} & \textbf{Modality} & \textbf{Acc5$\uparrow$} & \textbf{Acc2$\uparrow$} & \textbf{F1$\uparrow$} \\
    \midrule

    \multirow{5}{*}{Orig. Samples}
      & v  & 39.17 & 79.87 & 79.79 \\
      & a  & 41.36 & 78.34 & 74.90 \\
      & tv & \textbf{54.70} & \textbf{89.28} & \textbf{89.15} \\
      & ta & 45.95 & 80.31 & 78.64 \\
      & va & \underline{53.83} & \underline{86.43} & \underline{85.84} \\
    \midrule

    \multirow{5}{*}{Aug. Samples}
      & v  & 40.70 & 81.18 & 79.75 \\
      & a  & 42.67 & 78.77 & 77.28 \\
      & tv & \textbf{52.95} & \textbf{85.78} & \textbf{84.94} \\
      & ta & 46.83 & 80.74 & 79.04 \\
      & va & \underline{51.20} & \underline{85.34} & \underline{84.94} \\
    \midrule

    \multirow{5}{*}{Mixed (w/o QA)}
      & v  & 41.58 & 79.87 & 80.34 \\
      & a  & 43.11 & 77.68 & 77.64 \\
      & tv & \textbf{58.42} & \textbf{89.28} & \textbf{89.33} \\
      & ta & 50.55 & 82.28 & 82.11 \\
      & va & \underline{52.52} & \underline{85.56} & \underline{85.64} \\
    \midrule

    \multirow{5}{*}{Mixed (w/ QA)}
      & v  & 42.15 & 80.51 & 80.85 \\
      & a  & 43.50 & 80.27 & 80.24 \\
      & tv & \textbf{59.96} & \textbf{90.02} & \textbf{90.06} \\
      & ta & 51.86 & 84.25 & 83.84 \\
      & va & \underline{54.49} & \underline{88.62} & \underline{88.55} \\
    \bottomrule
  \end{tabular}
\end{table}

\begin{table}[t]
\centering
\caption{Ablation on negative sample construction strategies of QA scorer (CH-SIMS). ``Full'' denotes all three strategies.}
\label{tab:negative-strategies}
\small
\setlength{\tabcolsep}{5pt}
\renewcommand{\arraystretch}{1.15}
\begin{tabular}{@{}l ccccc@{}}
\toprule
\textbf{Strategy} & Acc5$\uparrow$ & Acc2$\uparrow$ & F1$\uparrow$ & MAE$\downarrow$ & Corr$\uparrow$ \\
\midrule
Feature Mixing only       & 61.27          & 88.40          & 88.54          & 0.2429          & 0.8265          \\
Random Masking only       & 51.86          & 84.25          & 84.08          & 0.3234          & 0.7200          \\
Label Flipping only       & 58.64          & 88.18          & 87.74          & 0.2455          & \textbf{0.8374}          \\
\midrule
\rowcolor{besthl}
Full          & \textbf{61.49} & \textbf{90.15} & \textbf{90.18} & \textbf{0.2400} & 0.8320 \\
\bottomrule
\end{tabular}
\end{table}

\begin{table}[t]
\centering
\caption{Ablation of weight mapping functions (CH-SIMS), with only the score-to-weight mapping varying. Linear corresponds to our default method.}
\label{tab:weight-mapping}
\small
\setlength{\tabcolsep}{5pt}
\renewcommand{\arraystretch}{1.15}
\begin{tabular}{@{}l ccccc@{}}
\toprule
\textbf{Mapping Function} & Acc5$\uparrow$ & Acc2$\uparrow$ & F1$\uparrow$ & MAE$\downarrow$ & Corr$\uparrow$ \\
\midrule
Uniform (no re-weighting) & 61.05          & 89.28          & 89.31          & 0.2372          & 0.8444 \\
Power ($\gamma$=0.5)      & 62.58          & 89.06          & 89.20          & 0.2389          & 0.8294          \\
Power ($\gamma$=2.0)      & 59.08          & 88.62          & 88.55          & 0.2429          & 0.8244          \\
Power ($\gamma$=4.0)      & 51.86          & 84.25          & 83.84          & 0.3072          & 0.7377          \\
Linear (End-to-End)       & 59.96          & 87.53          & 87.59          & 0.2346          & \textbf{0.8507}          \\
\midrule
\rowcolor{besthl}
Linear (ours)          & \textbf{61.49} & \textbf{90.15} & \textbf{90.18} & \textbf{0.2400} & 0.8320          \\
\bottomrule
\end{tabular}
\end{table}

\subsection{QA Scorer Analysis}
We evaluate the QA scorer design on the CH-SIMS dataset by comparing negative sample strategies and score-to-weight mapping functions. All experiments share the same random seeds, data splits, and hyperparameters as the main experiments.

\subsubsection{Negative Sample Strategies}
Table \ref{tab:negative-strategies} shows the performance of three negative sample strategies—cross-modal mismatch, feature corruption, and polarity flip—tested both individually and in combination. Using all three strategies together yields the best results, with Acc5 of 61.49\%, Acc2 of 90.15\%, and F1 of 90.18\%. Even the feature mixing strategy alone produces competitive results (Acc5 61.27\%, Acc2 88.40\%), showing that it captures the most common audiovisual misalignment issues in diffusion generation. The label flipping strategy gives moderate gains (Acc5 58.64\%), while random masking alone performs worst (Acc5 51.86\%). These results suggest that no single strategy can fully cover the real failure modes of diffusion generation. Only the combination of all three strategies provides an accurate proxy for key issues such as cross-modal inconsistency, feature degradation, and loss of conditioning signals, allowing the QA scorer to learn robust decision boundaries.

\subsubsection{Design Choices}
Table \ref{tab:weight-mapping} compares five different score-to-weight mapping functions while keeping the QA scores, data splits, negative sample strategies, and training hyperparameters fixed. The uniform mapping (w=1) serves as the baseline and achieves Acc5 of 61.05\%. Our default linear mapping improves Acc5 to 61.49\% and Acc2 to 90.15\%, giving the best overall performance. Power mapping with $\gamma=0.5$ performs similarly to the linear mapping (Acc5 62.58\%), but performance drops sharply as $\gamma$ increases to 2.0 and 4.0 (Acc5 59.08\% and 51.86\%, respectively). This pattern shows that excessively large $\gamma$ values sharpen the weight distribution too much, assigning near-zero weights to many useful but imperfect samples and thereby reducing the effective training data. Moderate linear or smooth weighting, by contrast, suppresses extreme noise while preserving the utility of the generated semantic diversity and yields the most stable training.
We also compare decoupled and end-to-end training. The decoupled approach—training the QA scorer independently in Stage 0 and freezing it during Stages 1 and 2—outperforms end-to-end training across all metrics. Acc2 rises from 87.53\% to 90.15\% and Acc5 from 59.96\% to 61.49\%, confirming that the decoupled design avoids gradient conflicts and maintains stable feature representations in the backbone network.

\subsection{Data Efficiency}
Table \ref{tab:qa_ratio} reports the performance of QASA on the CH-SIMS dataset when trained with different proportions of labeled data (10\%, 50\%, and 100\%). In all experiments, both the original and diffusion-augmented samples are reduced proportionally to maintain a consistent correspondence between datasets.

The results show that under extreme data scarcity with only 10\% of labeled samples, QASA achieves an Acc2 of 85.34\%, slightly outperforming the baseline trained on 100\% of the original data (Acc2 85.10\%). However, Acc5 (50.11\%), F1 (84.53\%), MAE (0.3650), and Corr (0.7473) remain below the full-data performance. This pattern matches the demands of fine-grained sentiment classification: Acc5 requires distinguishing subtle boundaries such as weak negative versus negative, which needs larger datasets to define complex decision surfaces. In contrast, Acc2, which captures coarse-grained positive/negative polarity, can quickly learn the core sentiment cues even with only 10\% of the data, thanks to the debiased, multi-style, and multi-timbre samples from QASA; it therefore exceeds the generalization of the full original dataset.

As the labeled proportion increases to 50\% and 100\%, all metrics improve steadily and substantially, with Acc2 reaching 89.28\% and 90.15\%, respectively. These gains indicate that the semantic diversity from diffusion-based augmentation together with QA scorer quality control effectively compensates for limited labeled data, producing rapid improvements in coarse-grained sentiment judgment under low-resource conditions and further strengthening fine-grained classification when more data is available. The results are fully consistent with the optimal performance observed in the main experiments under 100\% data and confirm the robustness and practical value of the framework in data-constrained scenarios.

\begin{table}[t]
\centering
\caption{Data efficiency analysis on CH-SIMS under different label ratios; \textbf{Orig.} indicate training on original samples; \textbf{QA.} indicate training on mixed samples(w/QA)}
\label{tab:qa_ratio}
\small
\setlength{\tabcolsep}{4.5pt}
\renewcommand{\arraystretch}{1.15}
\begin{tabularx}{\columnwidth}{@{}X c c c c c@{}}
\toprule
\textbf{Methods} 
& Acc5$\uparrow$ & Acc2$\uparrow$ & F1$\uparrow$ & MAE$\downarrow$ & Corr$\uparrow$ \\
\midrule
Orig.(100\%) & 52.10 & 85.10 & 85.00 & 0.3270 & 0.7490 \\
QA.(10\%)      & 50.11 & 85.34 & 84.53 & 0.3650 & 0.7473 \\
QA.(50\%)      & 60.83 & 89.28 & 89.39 & 0.2604 & 0.8147 \\
\midrule
\rowcolor{besthl}
QA.(100\%)     & \textbf{61.49} & \textbf{90.15} & \textbf{90.18} & \textbf{0.2398} & \textbf{0.8319} \\
\bottomrule
\end{tabularx}
\end{table}

\begin{figure}[t]
\centering
\includegraphics[width=\linewidth]{fig3.pdf}
\caption{t-SNE visualization of sentiment representations under different training configurations. }
\label{fig:tsne}
\end{figure}

\subsection{Visualization and Error Analysis}
We visualize sentiment representations under different training configurations with t-SNE. As shown in Figure \ref{fig:tsne}, training with only the original samples (a) produces substantial overlap among the negative (blue), neutral (gray), and positive (red) clusters, blurring decision boundaries. When only diffusion-augmented samples are used (b), the augmented samples (×) form a highly dispersed distribution clearly separated from the original samples (O). In mixed training without weighting (Mixed w/o QA) (c), the augmented samples scatter into isolated noise regions and disrupt the clustering structure of the original distribution. In weighted mixed training with the QA scorer (Mixed w/ QA) (d), however, the augmented samples (×) integrate smoothly into the original clusters and produce clear, continuous boundaries among the three sentiment classes. This progression from separation to noise to smooth integration highlights the QA scorer's central role in feature debiasing: it suppresses low-quality diffusion samples while allowing high-fidelity augmented samples to improve the original distribution.

\begin{figure}[t]
\centering
\includegraphics[width=\linewidth]{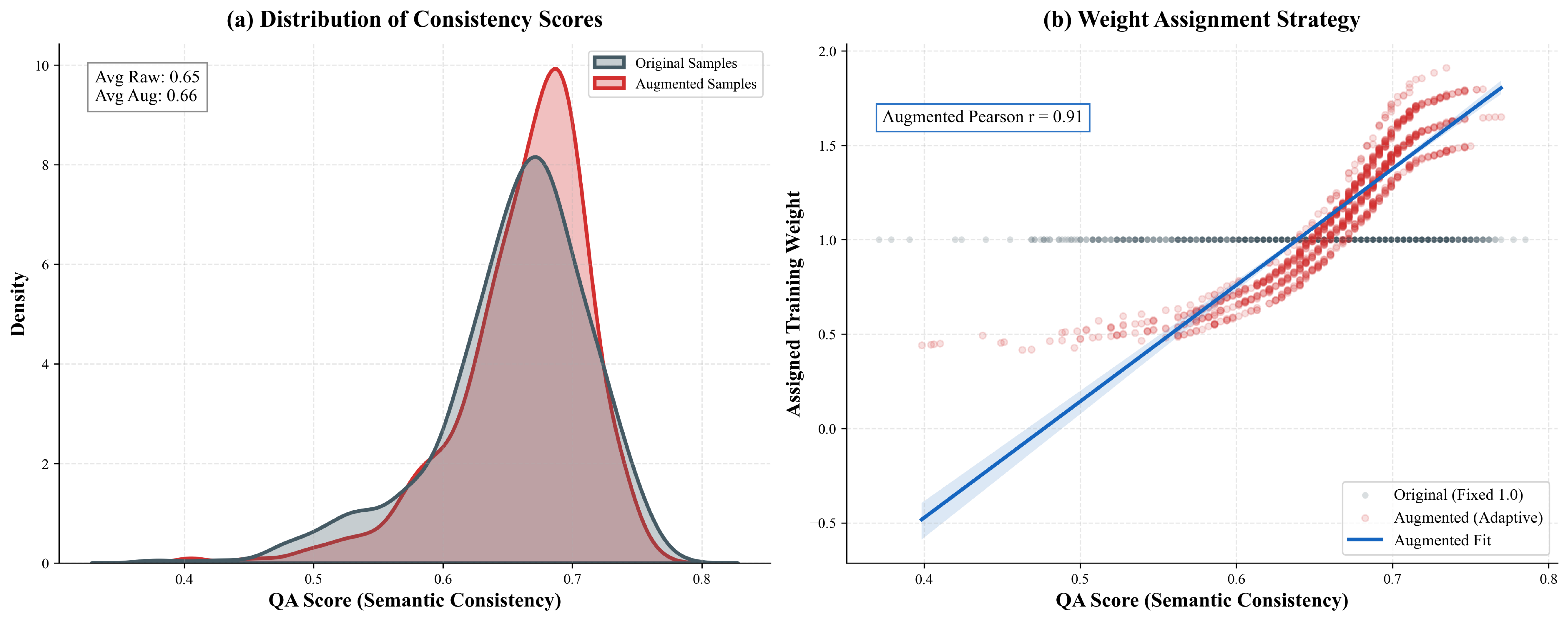}
\caption{(a) Distribution of QA consistency scores for original and augmented samples. (b) Mapping from QA scores to assigned training weights (Pearson $r=0.91$).}
\label{fig:qa-score}
\end{figure}

Figure \ref{fig:qa-score} provides quantitative support from the QA scorer. Panel (a) shows that the QA score distributions for original and augmented samples largely overlap, with mean scores of 0.65 and 0.66 respectively, indicating that diffusion-generated samples largely preserve semantic consistency. Panel (b) shows the mapping from QA scores to training weights, with a Pearson correlation coefficient of $r=0.91$. High-fidelity augmented samples (scores $>0.7$) receive high weights close to 1.5, while low-fidelity samples (scores $<0.5$) are strongly down-weighted below 0.5. This adaptive weighting ensures that the model focuses on reliable signals during training.

Despite QASA's clear performance gains, limitations remain. In challenging cases like extreme sarcasm or videos with low illumination and heavy background noise, diffusion augmentation still causes minor micro-expression loss or slight lip-sync deviations. While visual style transfer and timbre conversion effectively remove identity biases, extreme temporal alignment problems remain unsolved. Future work should explore more precise temporal consistency constraints or end-to-end lip-sync calibration modules.

Computational cost. The full three-stage training takes 15,504 seconds (about 4.31 hours), with Stage 0, Stage 1, and Stage 2 requiring 6,815, 4,900, and 3,789 seconds respectively. On two 80GB A800 GPUs, peak memory is 60.64GB and average GPU utilization is 39\%--40\%. Generating each augmented video sample takes roughly 2 minutes (80GB A800 GPU, peak memory 70.53GB) and each audio sample takes roughly 30 seconds (24GB 3090 GPU, peak memory 12GB). All costs are automatically logged for reproducibility.

\begin{figure}[t]
\centering
\includegraphics[width=\linewidth]{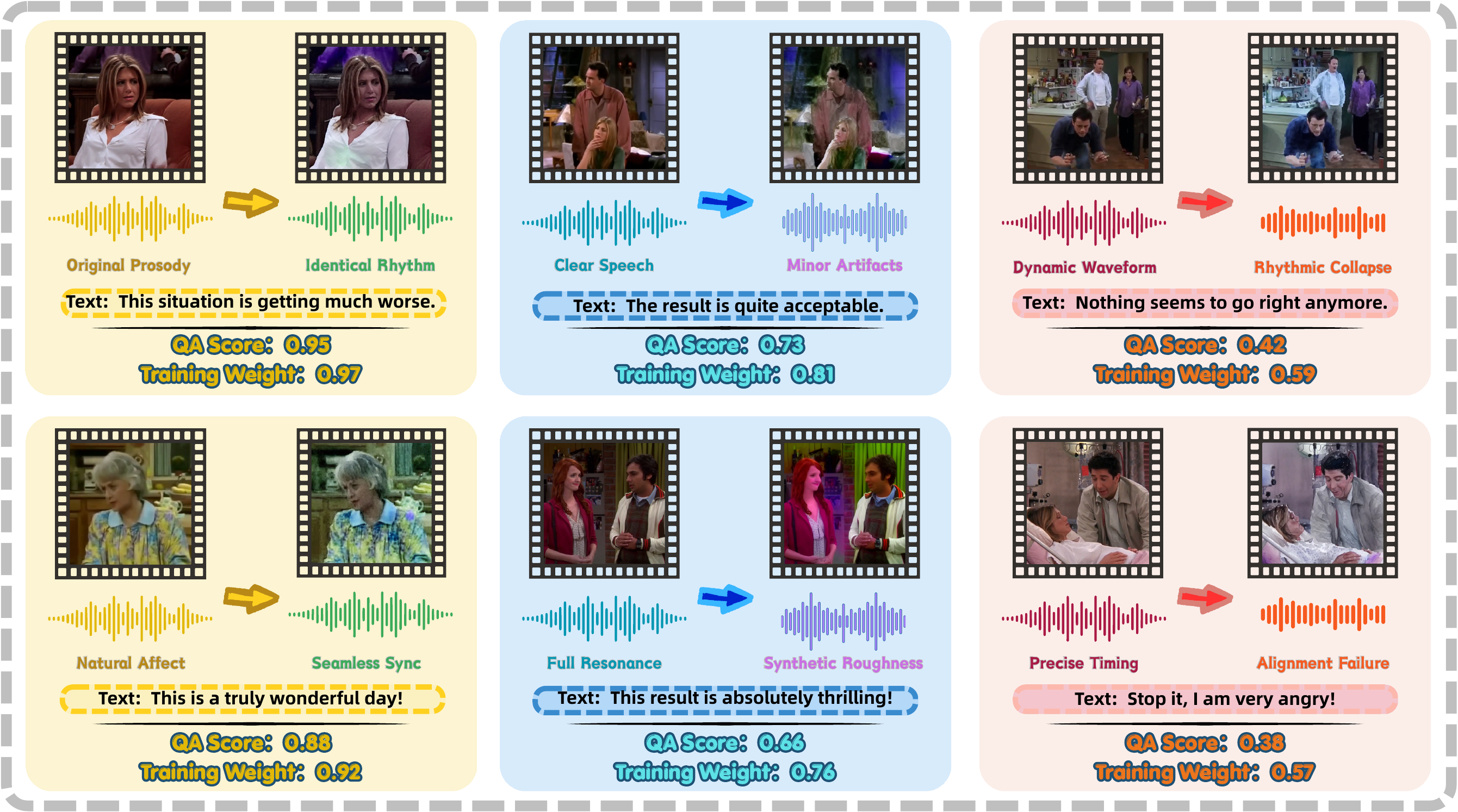}
\caption{Visualization of augmented samples.}
\label{fig:case_analysis}
\end{figure}

\subsection{Case Analysis}
We further evaluate the effectiveness of QASA by visualizing representative augmented samples in Figure~\ref{fig:case_analysis}. The columns are grouped according to QA Scores: high (yellow), medium (blue), and low (red). The left and right panels show the original and augmented video frames and audio waveforms, respectively. Each sample's QA Score and corresponding training weight are indicated below. High-scoring samples accurately preserve facial expressions and affective prosody, resulting in elevated training weights, while low-scoring samples receive lower weights. Medium-scoring samples are assigned intermediate weights, providing moderate augmentation during training without compromising model stability. These visualizations demonstrate that QASA effectively quantifies sample reliability and selectively incorporates high-quality augmentations to support robust cross-modal sentiment learning.

\section{Conclusion}
This paper presents QASA (Quality-Aware Semantic Augmentation), a novel data augmentation framework for multimodal sentiment analysis addressing high-quality data scarcity. The framework integrates diffusion-based style and timbre augmentation with a decoupled QA scorer, introducing targeted semantic variations while adaptively controlling training weights of augmented samples based on evaluated quality. This effectively enhances the stability and generalization of multimodal sentiment representations. Experiments demonstrate QASA consistently outperforms existing methods on the CH-SIMS, CMU-MOSI, and MUStARD benchmarks, exhibiting strong robustness in cross-modal conflict scenarios.

\bibliographystyle{ACM-Reference-Format}
\bibliography{sample-base}

\end{document}